\documentclass[letterpaper, 10 pt, conference]{ieeeconf}

\usepackage{graphicx}
\usepackage{float}
\usepackage{cite}
\usepackage{xcolor}
\usepackage{siunitx}
\usepackage{amsmath}
\usepackage{makecell}
\newcommand{\robotname}{{{PuffyBot}}}

\usepackage[hidelinks]{hyperref}

\title{\robotname: An Untethered Shape-Morphing Robot for Multi-Environment Locomotion}

\author{Shashwat Singh, Zilin Si, and Zeynep Temel\\
\small Robotics Institute, Carnegie Mellon University, Pittsburgh, PA 15213, USA
}
    
\begin{document}
\title{\robotname: An Untethered Shape Morphing Robot for Multi-environment Locomotion}

\maketitle
\begin{abstract}
Amphibians adapt their morphologies and motions to accommodate movement in both terrestrial and aquatic environments. Inspired by these biological features, we present \robotname, an untethered shape morphing robot capable of changing its body morphology to navigate multiple environments. Our robot design leverages a scissor-lift mechanism driven by a linear actuator as its primary structure to achieve shape morphing. The transformation enables a volume change from \SI{255.00}{\centi\meter\cubed} to \SI{423.75}{\centi\meter\cubed}, modulating the buoyant force to counteract a downward force of \SI{3.237}{\newton} due to \SI{330}{\gram} mass of the robot. A bell-crank linkage is integrated with the scissor-lift mechanism, which adjusts the servo-actuated limbs by \SI{90}{\degree}, allowing a seamless transition between crawling and swimming modes. The robot is fully waterproof, using thermoplastic polyurethane (TPU) fabric to ensure functionality in aquatic environments. The robot can operate untethered for two hours with an onboard battery of \SI{1000}{\milli\ampere\hour}. Our experimental results demonstrate multi-environment locomotion, including crawling on the land, crawling on the underwater floor, swimming on the water surface, and bimodal buoyancy adjustment to submerge underwater or resurface. These findings show the potential of shape morphing to create versatile and energy efficient robotic platforms suitable for diverse environments.
\end{abstract}


\IEEEpeerreviewmaketitle
\section{Introduction}
Locomotion across different environments, particularly the transition between land and water, presents unique challenges that require biomechanical adaptations. Amphibians, such as turtles and frogs, exhibit remarkable adaptability in their locomotion, enabling them to efficiently navigate both terrestrial and aquatic environments~\cite{glasheen1996hydrodynamic,parker1922crawling,duellman1994biology}. Their ability to switch between different modes of movement such as crawling, walking, jumping, and swimming allows them to exploit diverse habitats to find food, lay eggs, escape predators, or capture prey. Turtles, for example, employ a slow and deliberate gait on land, using their sturdy limbs to crawl, while their webbed feet or flippers facilitate swimming through a rowing or flapping motion in water~\cite{mayerl2017novel}. Similarly, frogs utilize powerful hindlimbs to jump on land, whereas in water they rely on synchronized leg strokes to propel themselves forward~\cite{nauwelaerts2005swimming}. In addition, some amphibians and aquatic species transition from the surface of water to underwater environments by regulating buoyancy through lung inflation~\cite{fossette2010behaviour} or by using the swimming bladder~\cite{alexander1982buoyancy}. 
These biomechanical adaptations~\cite{trimmer2014bone} play a critical role in facilitating the challenges and strategies associated with multi-environment locomotion.

Inspired by such biological mechanisms, researchers have developed multi-environmental robots capable of transitioning between different environments~\cite{ren2021research,sun2023embedded,patel2023highly,sihite2023multi,dudek2007aqua}. These robots, designed to navigate complex unstructured terrains~\cite{ramirez2023multimodal}, have significant applications in search and rescue~\cite{lindqvist2022multimodality}, and environmental exploration, where adaptability is crucial for efficient mobility~\cite{ijspeert2020amphibious}. 

\begin{figure}[t]
  \centering
  \includegraphics[width=\linewidth]{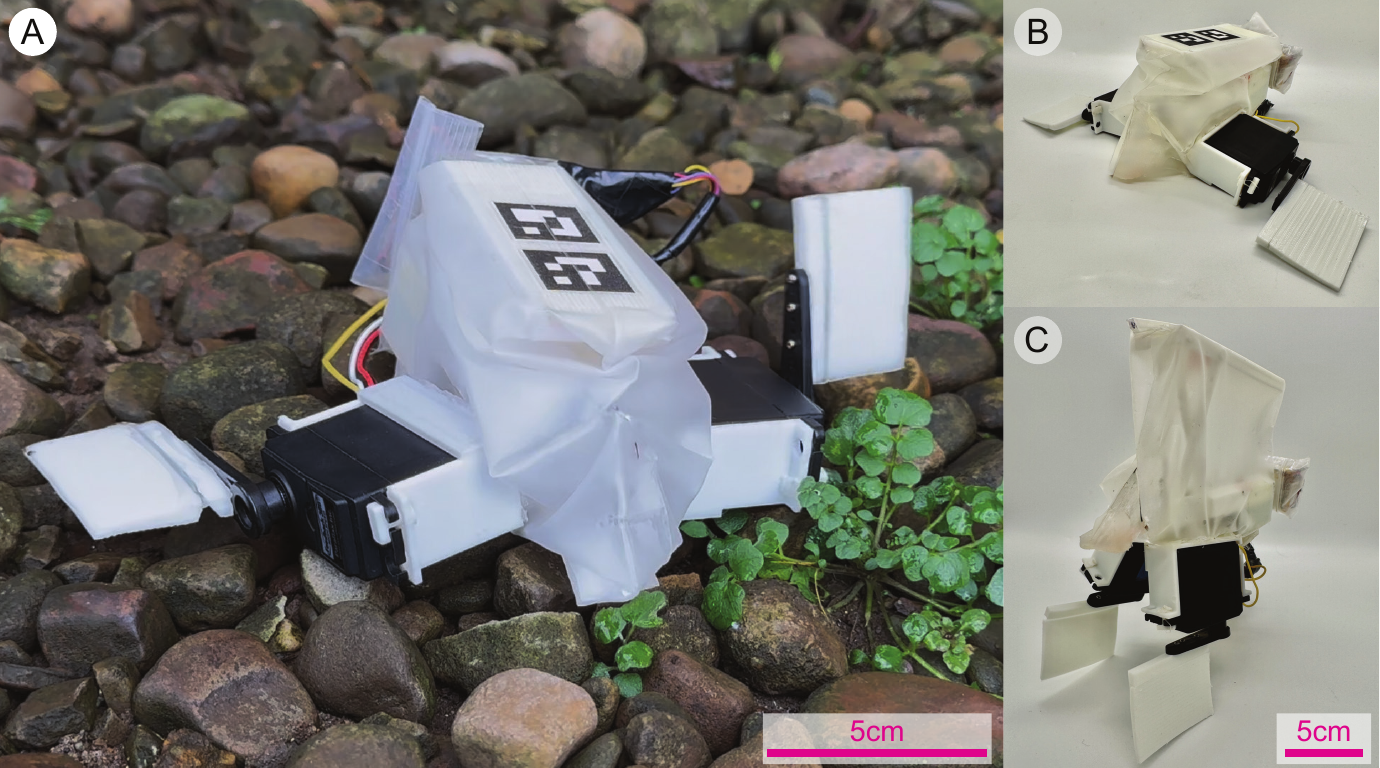}
  \caption{(A) PuffyBot photographed in natural terrain, equipped with onboard actuation, control electronics, and power supply. 
  (B) Crawling mode: the scissor-lift body is fully compressed and the fins lie flat, providing stability and traction for crawling on land or along the underwater floor. 
  (C) Swimming mode: the scissor-lift body expands and the fins rotate downward to align with the body, forming a streamlined configuration for surface swimming.}
  \label{fig:robot}
  \vspace{-1.0em}
\end{figure}

Performing multi-environment locomotion in robots typically involves a combination of shape morphing and gait transitions~\cite{sun2023embedded,baines2022multi} or employing various propulsion and actuation mechanisms~\cite{sihite2023multi,hwang2022shape,crespi2013salamandra, chen2017biologically, chen2018controllable}. One common approach to building shape morphing multi-environment robots is the utilization of soft materials integrated with heating coils or shape memory alloys (SMA). Sun et al.~\cite{sun2023embedded} demonstrated an amphibious robot that features embedded shape actuation, locking, and sensing capabilities in its limbs, employing shape memory polymer (SMP) and joule heating via a twisted coil actuator for amphibious locomotion. Similarly, the Amphibious Robotic Turtle (ART) developed by Baines et al.~\cite{baines2022multi} incorporated a shape morphing limb capable of transitioning between a leg and a flipper configuration, using dielectric elastomer actuators (DEA) through joule heating, for amphibious movement. Patel et al.~\cite{patel2023highly} introduced a bi-stable reconfigurable amphibious robot that uses a SMA coil and silicone elastomer to morph the body and limb. Although these techniques have demonstrated remarkable potential for reconfigurable robots, their morphing processes are often constrained by high energy consumption and long transformation times as a result of reliance on joule heating. These systems typically require between \SI{20}{\joule} and \SI{3000}{\joule} of energy for shape transformation and consume approximately \SI{20}{\watt} to \SI{60}{\watt} of power during normal operation. Alternatively, to meet these power requirements, existing shape-morphing robots often face limitations such as tethered operation or the need for a heavy onboard battery for long range operation~\cite{sun2023embedded, baines2022multi}.

Another class of multi-environment robots achieves locomotion through mechanical architectures specifically designed for operation across different environments. Salamandra Robotica~\cite{crespi2013salamandra}, for instance, utilizes a modular structure consisting of an actuated spine and four articulated limbs, enabling it to walk on land and swim using anguilliform motion in water. However, the robot has considerably large size that measures more than \SI{100}{\centi\meter} in length, \SI{4.7}{\centi\meter} in width, and \SI{5.8}{\centi\meter} in height along with its segmented body components required for multi-environment functionality, constrain its ability to maneuver in confined or cluttered environments. At the microscale, some multi-environment robots leverage buoyancy adaptation by adjusting their size and weight. Chen et al.~\cite{chen2017biologically} developed a \SI{175}{\milli\gram} microrobot capable of flying and swimming on the basis of flapping, and employing an electrolytic buoyancy chamber for underwater transitions. Another microrobot~\cite{chen2018controllable} weighing \SI{1.6}{\gram} uses feet with electrowetting pads (EWP) and passive flaps to walk on the surface of the land and water and sink underwater by breaking the surface tension. To break the surface tension and sink underwater, it needs \SI{600}{\volt} supplied to its feet. In both cases, the robots must weigh less than a few grams and rely on buoyancy adaptation techniques such as electrowetting and electrolysis, which require high voltages, limit scalability, and often require tethered operation.

\begin{table}[t]
\centering
\caption{Comparison of shape morphing amphibious robots}
\label{tab:morphing_comparison}
\renewcommand{\arraystretch}{1.2}
\begin{tabular}{lcccc}
\hline
\textbf{Robot}  & \textbf{Morphing Energy} & \textbf{Environment} \\
\hline
ART~\cite{baines2022multi} & $\sim$3000~J & Land / Water \\
SMART~\cite{sun2023embedded} & $\sim$1400~J & Land / Water \\
Amphibious Bot~\cite{patel2023highly} & $\sim$22~J & Land / Water \\
\textbf{PuffyBot (This Work)} & \textbf{$\sim$7~J} & \makecell{\textbf{Land / Water}\\ \textbf{Underwater}} \\
\hline
\end{tabular}
\vspace{-1.0em}
\end{table}

In this paper, we present \robotname, an untethered multi-environment shape morphing robot (Figure~\ref{fig:robot}.A) capable of crawling on land, swimming in water, crawling along the underwater floor and seamlessly transitioning between these modes (See supplementary video). The robot weighs \SI{330}{\gram} and has a compact form factor, with a length of \SI{9}{\centi\meter}, a width of \SI{4.5}{\centi\meter}, and a height that varies between \SI{4}{\centi\meter} and \SI{9}{\centi\meter}. Our design incorporates a bimodal buoyancy control mechanism that enables controlled submerging and resurfacing, drawing inspiration from Archimedes principles. The complete transition from staying afloat to sinking, and then resurfacing, requires less than \SI{7}{\joule} of energy, demonstrating the high efficiency of the shape morphing and buoyancy adapting system (Table~\ref{tab:morphing_comparison}). Powered by a \SI{3.7}{\volt}, \SI{1000}{\milli\ampere\hour} battery, the robot is capable of continuous operation for up to two hours. Additionally, the robot's structure is 3D printed, which simplifies fabrication, enabling rapid and accessible prototyping. Furthermore, we evaluated the locomotion performance of the robot in different environments, measuring an average locomotion speed of \SI{0.70}{\centi\meter\per\second} on land, \SI{0.75}{\centi\meter\per\second} for swimming, and \SI{0.24}{\centi\meter\per\second} for crawling on the underwater floor (See supplementary video). Our approach not only enhances transition efficiency and reduces power consumption, but also has potential applicability of amphibious robots in real-world exploration from long-term shallow water inspection to multi-environment monitoring.


\section{Design and Fabrication}

We design our robot based on the Archimedes principle and coupling mechanisms to enable change in buoyancy and multi-environment locomotion. The robot's main body is structured with a scissor-lift and is controlled by a linear actuator to change its shape and volume and further adjust its buoyancy. On both sides of the robot body, two continuous feedback control servo motors are coupled to the scissor-lift arm using the bell crank linkage. The coupled linkage changes the orientation of the robot's fins and their morphology for effective locomotion on the land and in the water. During crawling, the fins remain flat to the surface as shown in Figure~\ref{fig:robot}.B to keep the robot's center of mass (CoM) low to the ground and to better push the robot forward by getting more traction from the surface. During swimming, the fins transform into a vertical position Figure~\ref{fig:robot}.C, which is more streamlined to the body to achieve better stability and less drag. The robot body is waterproofed with TPU fabric~\cite{fabric} to prevent leakage during underwater operation and to enable buoyancy adjustment. Furthermore, 3D printed compliant fins are used to aid in walking and swimming. Finally, a custom PCB is designed to control the robot over Wi-Fi and to power the actuators.

\begin{figure}
  \centering
  \includegraphics[width=0.9\linewidth]{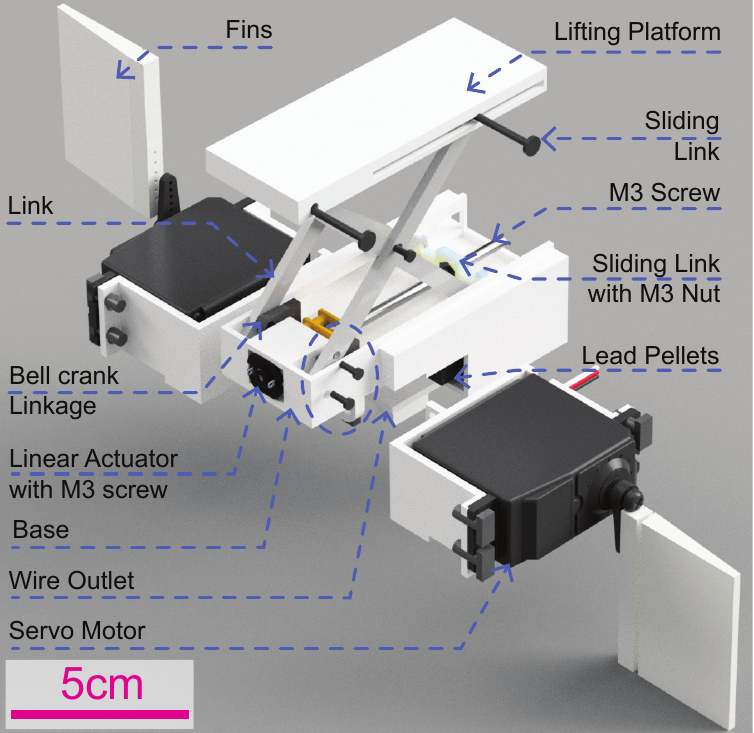}
  \caption{An exploded view of the CAD model shows the robot design including links, mechanisms and actuators.}
  \label{fig:exploded}
  \vspace{-1em}
\end{figure}

\subsection{Shape morphing body}

We use a scissor-lift for designing the shape morphing robot body. The detailed links and mechanisms of the robot are shown in the exploded CAD view (Figure~\ref{fig:exploded}). The base of the scissor-lift remains stationary, and the lifting platform can move vertically to change the height and further the volume of the robot.  On the base, we mount a linear actuator horizontally with a M3 \SI{55}{\milli\meter} screw~\cite{Screw_M3_motor}. The M3 screw is then connected to a 3D printed sliding link with M3 nut embedded in it using super glue~\cite{super_glue} and friction fit. The front two links of the scissor-lift are coupled to bell crank linkages to change the orientation of the servo motors. A pocket is included at the base to add lead pellets to increase the mass of the robot which aids the robot to sink in the compressed state. They are also used to balance the weight of the front servo motors and adjust the center of mass (CoM) of the robot to increase stability.



\subsection{Mode change using coupled bell-crank linkage}

\begin{figure}[t]
  \centering
  \includegraphics[width=0.8\linewidth]{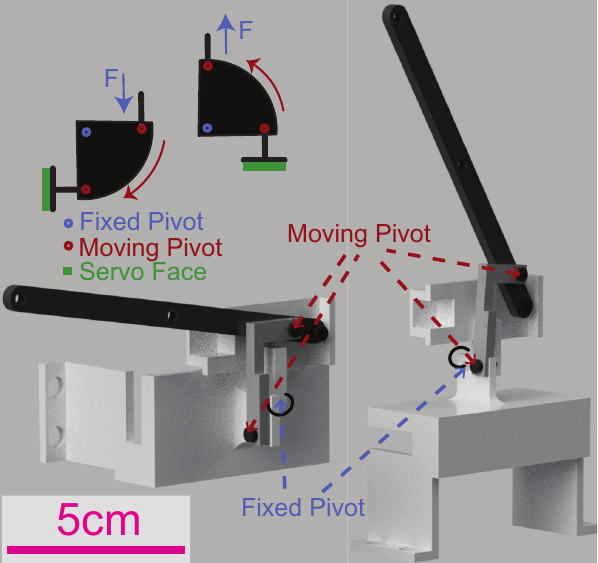}
  \caption{Bell-crank linkage coupling the scissor-lift link and servo mount. The mechanism converts vertical motion into a \SI{90}{\degree} servo reorientation, enabling transitions between horizontal (crawling) and vertical (swimming) mode.}
  \label{fig:bell_crank}
  \vspace{-1em}
\end{figure}

The servo motor mount is mechanically coupled to the scissor-lift link through a bell-crank mechanism that reorients the servo by approximately \SI{90}{\degree} between the robot’s two primary crawling and swimming locomotion modes. The working principle of this linkage is illustrated in Figure~\ref{fig:bell_crank}. An L-shaped lever is pivoted at its center (fixed pivot), converting the vertical motion of the scissor-lift link at the upper moving pivot (input - force) into a horizontal rotation at the lower moving pivot (output - servo face orientation). As the lift expands or contracts, this motion is transferred through the bell-crank to rotate the servo mount, thereby changing the fin orientation. The diagram highlights this transition with the servo face, illustrating the \SI{90}{\degree} change in orientation.

In the compressed state, the servos remain horizontal, lowering the center of mass and providing stability for crawling on land and along the underwater floor. In the expanded state, the bell-crank linkage rotates the servo mounts vertically, aligning the fins with the body to form a streamlined configuration suitable for swimming on the water surface.

\subsection{Soft compliant fin and gait design}

\begin{figure}[t]
  \centering
  \includegraphics[width=0.8\linewidth]{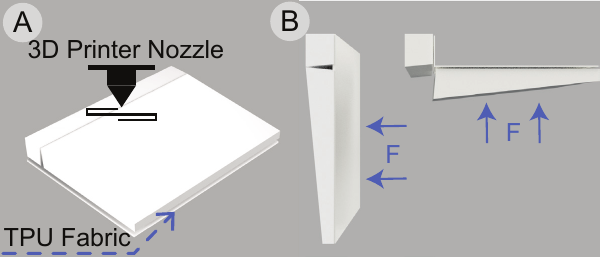}
  \caption{(A) 3D printing of a compliant fin directly onto TPU fabric using TPU filament. 
  (B) The soft compliant fin acts as a flexible joint: it self-locks during crawling and backward stroke to generate thrust, and bends upward during the forward recovery stroke to minimize drag in water.}
  \label{fig:fin}
  \vspace{-1em}
\end{figure}

The compliant fins are manufactured by directly 3D printing TPU filament onto a TPU fabric substrate (Figure~\ref{fig:fin}). A sheet of TPU fabric~\cite{fabric} is fixed to the print bed with Kapton tape, and a Bambu Lab X1 Carbon printer deposits molten TPU filament, forming a strong bond between layers through heat and inter-diffusion. The fabric serves as a flexible hinge that allows the fin to bend passively during motion.

When the fin moves backward, the joint self-locks against the ground or fluid, generating thrust. During the forward stroke, it bends with the flow, minimizing resistance. This passive asymmetry provides propulsion in both terrestrial and aquatic environments.

Crawling gait: In the compressed state, the fins remain flat against the surface to maximize traction. Alternating \(360^\circ\) rotations of the fins generate forward motion as each fin pushes against the ground while the other resets.

Swimming gait: In the expanded state, the fins oscillate through approximately \(180^\circ\) of motion with faster power strokes and slower recovery strokes, producing net forward thrust while minimizing backward slip.

The combination of compliant fin mechanics and asymmetric gait patterns enables the robot to adapt its propulsion to both land and water using the same actuation system.

\subsection{Electronics and control}

The robot's electronics are comprised of a custom printed circuit board (PCB) with a microcontroller module (Xiao ESP32-S3, Seeed Studio) for computation and remote control over Wi-Fi. The linear actuator (150:1 \SI{6}{\volt}, Walfront)~\cite{Screw_M3_motor} used to adjust the scissor-lift is powered through a \SI{10.8}{\volt}, \SI{2}{\ampere} dual H-bridge (DRV8833, Texas Instruments). Two GPIO pins of the custom PCB are connected to the Arduino Nano Every interrupt pin to control continuous servo motors (Parallax Feedback 360°~\cite{servo}). 

The robot carries its own power supply in the form of a lithium polymer battery. The battery is rated for \SI{3.7}{\volt} with a capacity of \SI{1000}{\milli\ampere\hour}. The battery connects to the board with a JST connector so that we can easily disconnect and recharge the battery. The battery has a discharge rate of up to 20C, which
meets the continuous power demands of the robot. During robot operation, the robot measures an average current of \SI{500}{\milli\ampere}, resulting in an operational run time of two hours. 

\subsection{Waterproofing}

Waterproofing is essential to protect onboard electronics and enable buoyancy control during underwater operation. The robot body is constructed from transparent soft thermoplastic polyurethane (TPU) fabric~\cite{fabric}, chosen for its flexibility, strength, and ability to be heat-sealed. The TPU sheet is folded and sealed along its edges using a heat sealer at \SI{125}{\celsius} to form a pouch into which the robot body is inserted. The openings for the servo joints and actuator wires are cut precisely, glued, and reinforced with epoxy to ensure a watertight seal. The servo motors are internally coated with a waterproofing spray~\cite{spray} to prevent corrosion, and the control electronics and battery are enclosed in a small sealed plastic pouch attached to the rear of the robot for easy removal for programming and recharging.

\section{Modeling and Design Space for Buoyancy Adaptation}

This section develops a model to describe \robotname’s shape morphing buoyancy adaptation and to construct a design space that generalizes its buoyancy control strategy. 
Together, these models answer two key questions: (i) can the robot reliably change its buoyancy through shape morphing, and (ii) how do its geometric and mass parameters influence the resulting buoyant force?

\subsection{Adaptive buoyancy with shape morphing body}

The weight and volume of the robot are critical considerations in its design, as any change in weight or volume can significantly impact the robot's buoyancy. Specifically, an increase in weight will generate excess downward force, while an increase in volume will produce additional upward buoyant force, potentially destabilizing the robot's adaptive buoyancy mechanism. Assuming the density of the water is approximately \SI{1}{\gram\per\cubic\centi\meter}, this implies that \SI{1}{\gram} of the mass will displace \SI{1}{\cubic\centi\meter} of the water. With the robot weighing \SI{330}{\gram} (equivalent to \SI{330}{\centi\cubic\meter} of displaced water), its net volume must be less than \SI{330}{\centi\cubic\meter} to achieve negative buoyancy and greater than \SI{330}{\centi\cubic\meter} to achieve positive buoyancy.  

The robot's total volume is comprised of three primary components: the robot body, electronics and battery pouch, and servo motors. In its compressed state, the robot has a volume of \SI{135}{\centi\cubic\meter}, while in its expanded state, the volume increases to \SI{303.75}{\centi\cubic\meter}. The servo motors and battery pouch contribute fixed volumes of \SI{72}{\centi\cubic\meter} and \SI{48}{\centi\cubic\meter}, respectively. This results in total volumes of \SI{255.00}{\centi\cubic\meter} in the compressed state and \SI{423.75}{\centi\cubic\meter} in the expanded state. When these values are compared with the critical volume of \SI{330}{\centi\cubic\meter}, it is evident that the robot remains less buoyant in the compressed state and more buoyant in the expanded state. Therefore, the robot can effectively transition between positive and negative buoyancy states on the basis of its configuration.

\begin{figure}
  \centering  \includegraphics[width=0.7\linewidth]{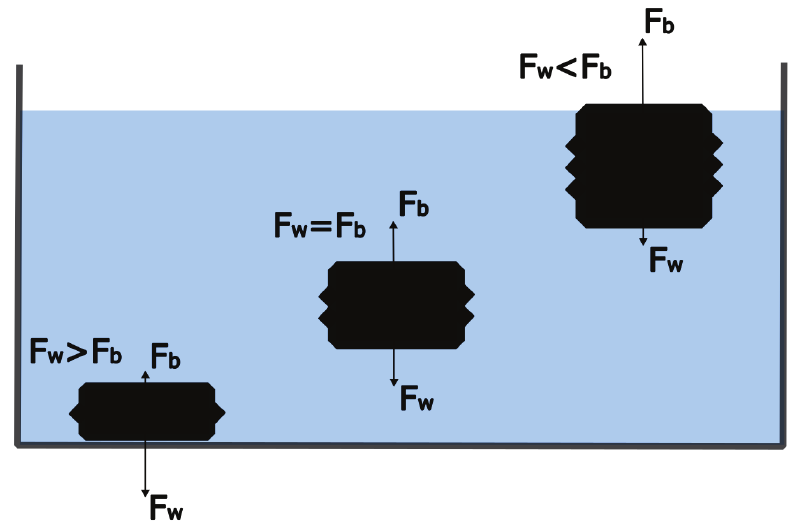}
  \caption{Free-body diagram showing three buoyancy states of the robot with change in volume: $F_\text{w} > F_\text{b}$: negative, $F_\text{w} = F_\text{b}$: neutral, and $F_\text{b} > F_\text{w}$: positive buoyancy.}
  \label{fig:fbd}
  \vspace{-1.0em}
\end{figure}

\subsection{Buoyancy Modeling}

Figure~\ref{fig:fbd} presents the free-body diagram of the robot in water in three states: when it rests at the bottom of the tank, when it achieves neutral buoyancy, and when it becomes positively buoyant and floats to the water surface. 

The downward force $F_{w}$ is caused by gravity and can be expressed as:
\vspace{-0.5em}
\begin{align}
F_{\text{w}} = m_{\text{robot}} \cdot g
\label{eq:weight}
\end{align}

This force remains constant and is determined by the robot's mass ($0.330 \, \text{kg}$) and gravitational acceleration ($g = 9.81 \, \text{m/s}^2$), yielding:
\[
F_{\text{w}} = 0.330 \cdot 9.81 = 3.237 \, \text{N}
\]

In contrast, the upward buoyant force $F_{b}$ depends on the total volume of water displaced by the robot, which includes both a variable component based on its height and a fixed internal volume ($V_{\text{fixed}} = 120 \, \text{cm}^3$) contributed by non-deformable parts such as the servo motors and battery pouch. Thus, the buoyant force is given by:
\vspace{-0.2em}
\begin{align}
F_{\text{b}} &= \rho_{\text{water}} \cdot (\Delta V_{\text{variable}} + V_{\text{fixed}}) \cdot g \label{eq:vol_total}
\end{align}

Although the enclosure is watertight, the morphing process behaves quasi-isobarically: the internal air pressure remains near ambient because small amounts of gas can diffuse through the TPU film and sealed interfaces, allowing slow pressure equalization during expansion and compression. As a result, the buoyant force can be modeled primarily as a function of displaced volume, without explicitly accounting for pressure variations inside the chamber.

The variable volume is defined by the robot height $h_{\text{robot}}$ and its cross-sectional area ($l_{\text{robot}} \cdot w_{\text{robot}} = 9 \cdot 4.5 = 40.5 \, \text{cm}^2$):
\vspace{-0.2em}
\begin{align}
\Delta V_{\text{variable}} &= l_{\text{robot}} \cdot w_{\text{robot}} \cdot h_{\text{robot}} \label{eq:height_volume}
\end{align}

To determine the height at which the robot reaches neutral buoyancy, we set $F_b = F_w$ and solve for $h_{\text{robot}}$:

\[
3.237 = 1000 \cdot \left( \frac{40.5 \cdot h_{\text{robot}} + 120}{10^6} \right) \cdot 9.81
\]

\[
\Rightarrow \left( \frac{40.5 \cdot h_{\text{robot}} + 120}{10^6} \right) = \frac{3.237}{1000 \cdot 9.81}
\]

\[
\Rightarrow 40.5 \cdot h_{\text{robot}} + 120 = 330 \Rightarrow h_{\text{robot}} = \frac{210}{40.5} \approx 5.2 \, \text{cm}
\]

Thus, in the compressed state, the robot is negatively buoyant and sinks to the bottom of the tank, while in the expanded state it displaces enough water to generate a net upward force and float to the surface. According to the model, neutral buoyancy occurs at a body height of approximately \SI{5.2}{\centi\meter}, which separates the negatively and positively buoyant regimes. 

These results characterize the fundamental relationship between robot geometry, mass, and buoyant force. Building on this analysis, the next section presents a design-space representation that generalizes these relationships across a wider range of body heights and masses to guide future design parameter selection.

\begin{figure}[!ht]
  \centering
  \includegraphics[width=1\linewidth]{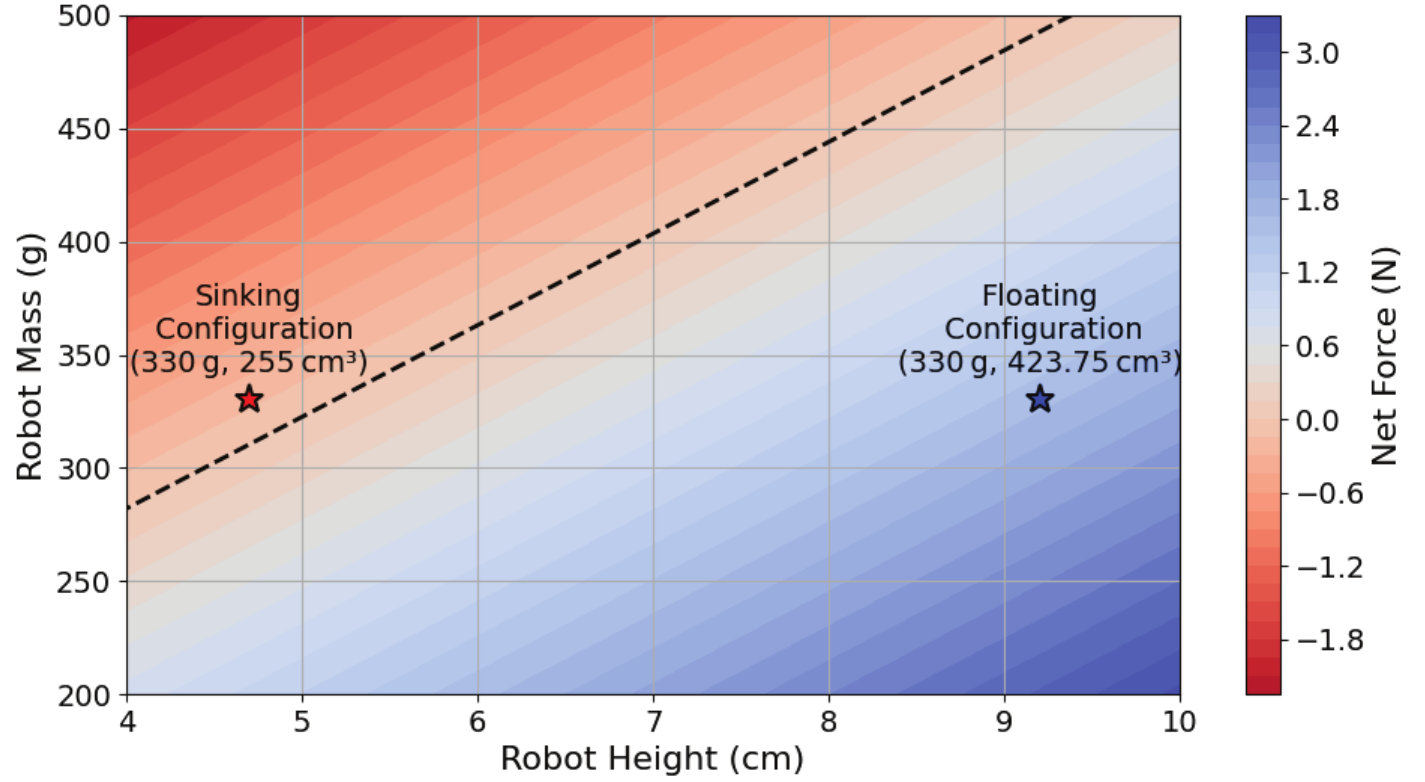}
  \caption{Design space showing net buoyant force as a function of robot height and mass. The black dashed line represents the neutral buoyancy threshold. Red and blue star markers denote the sinking (\SI{330}{\gram}, \SI{255.00}{\centi\meter\cubed}) and floating (\SI{330}{\gram}, \SI{423.75}{\centi\meter\cubed}) configurations.}
  \label{fig:designspace}
  \vspace{-1.0em}
\end{figure}

\subsection{Design Space Analysis for Buoyancy Adaptation}

To generalize the robot's buoyancy control strategy, we developed a design space model that relates the robot body height and mass to the net buoyant force in water. The total displaced volume includes both the variable body volume and a constant fixed volume of \SI{120}{\centi\meter\cubed} contributed by the servo motors and the battery pouch. The net force is computed by subtracting the gravitational weight from the upward buoyant force, based on Archimedes principle.

Figure~\ref{fig:designspace} presents the resulting contour map of the net buoyant force across a range of robot masses (\SI{200}{\gram} to \SI{500}{\gram}) and body heights (\SI{4}{\centi\meter} to \SI{10}{\centi\meter}). The black dashed line indicates the neutral buoyancy threshold, where the buoyant and gravitational forces are equal. Configurations below the line are positively buoyant (floating), while those above the line are negatively buoyant (sinking).

The robot configurations are overlaid as star markers. In the compressed state (\SI{4.5}{\centi\meter} height), the robot has a total volume of \SI{255.00}{\centi\meter\cubed} and is negatively buoyant. In the expanded state (\SI{9}{\centi\meter} height), the total volume increases to \SI{423.75}{\centi\meter\cubed}, allowing the robot to generate sufficient upward force to float. These transitions are visually confirmed in the plot, highlighting the system's ability to switch between sinking and floating modes via shape morphing.

\begin{figure}
  \centering  \includegraphics[width=\linewidth]{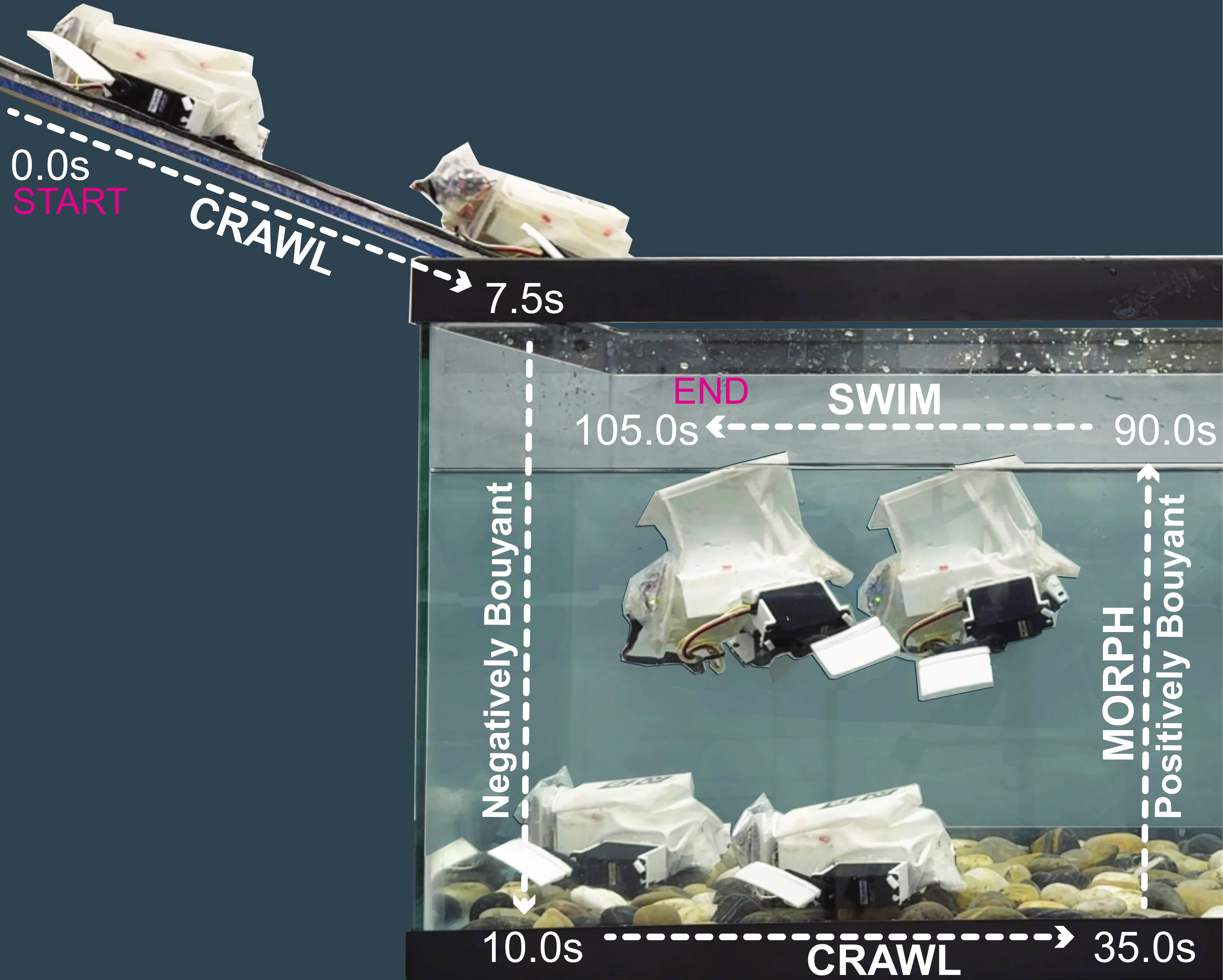}
  \caption{Multi-environment locomotion and transitions: The robot starts from a ramp at time \SI{0}{\second} and crawled down for \SI{7.5}{\second} and then drops in water and goes to the water surface at \SI{10}{\second}. The robot crawls underwater for \SI{25}{\second} for one body length. After crawling underwater, the robot expands its body to change its buoyancy to positive at time \SI{90}{\second} and swims backward and finally stops at \SI{105}{\second}.}
  \vspace{-1.0em}
  \label{fig:transition}
\end{figure}

\section{Experiments and Results}

\subsection{Experimental Setup}

\begin{figure*}[t]  
  \centering
  \includegraphics[width=0.8\linewidth]{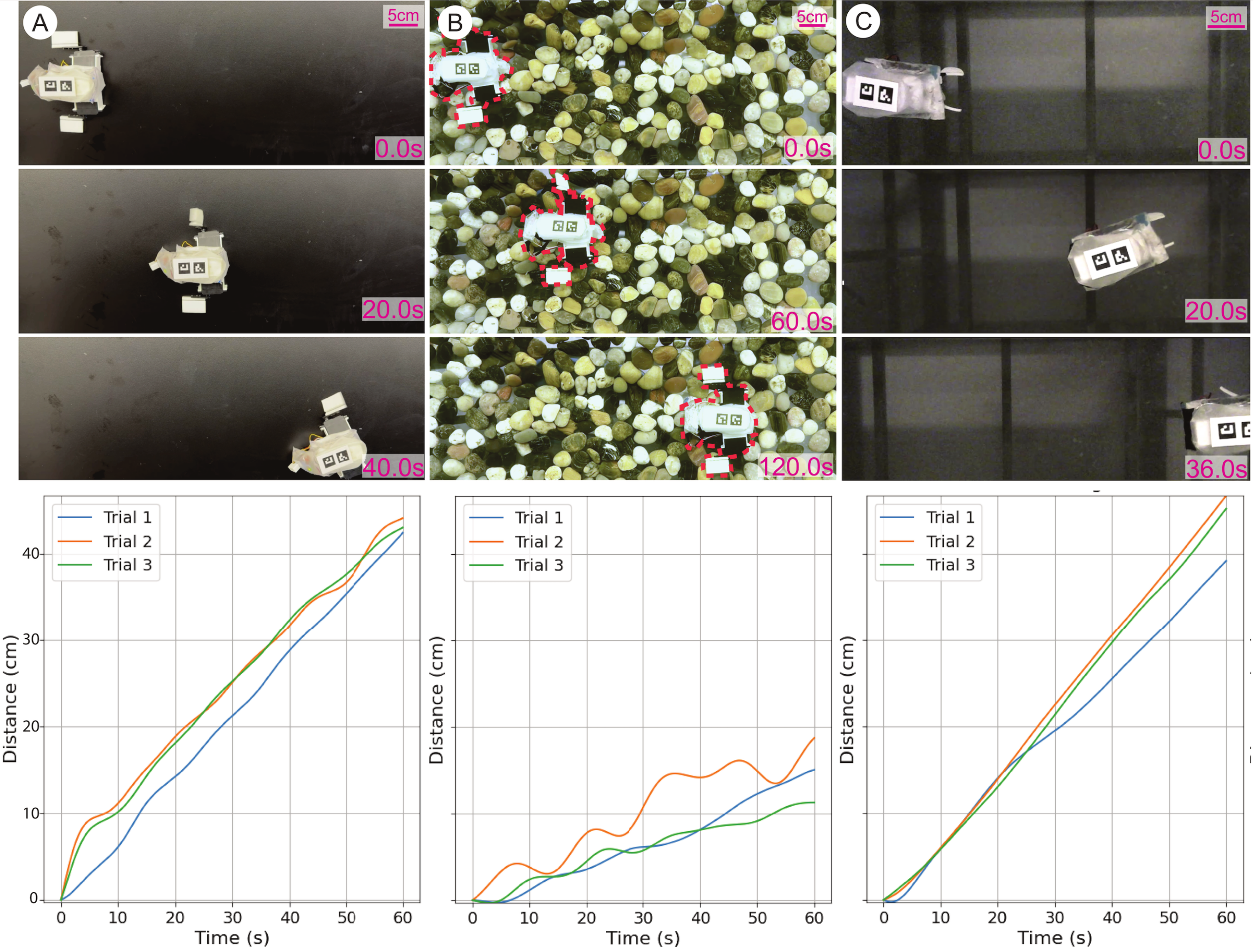}  
  \caption{Robot demonstrating versatile locomotion capabilities on land, underwater floor and on water surface with distance - time plot are shown below each panel: the robot performing locomotion in different environments and plotted for three different trials for \SI{60}{\second} below them. (A) crawling on a flat ground, (B) crawling on underwater floor over pebbles, and (C) swimming on the water surface.}  
  \label{fig:multi-environment} 
  \vspace{-1.0em}
\end{figure*}

To evaluate the crawling performance of the robot, experiments are conducted on a flat surface using open-loop control. Two ArUco markers are attached to the top surface of the robot to facilitate motion tracking as shown in Fig.~\ref{fig:multi-environment}. A camera is placed \SI{60}{\centi\meter} above the robot, recording from a top-down perspective at 30 frames per second (fps) for tracking purposes. Similarly, for swimming and underwater crawling tests, the robot is placed in a water tank measuring \SI{90}{\centi\meter} in length, \SI{45}{\centi\meter} in width, and \SI{42}{\centi\meter} in height with water filled to \SI{30}{\centi\meter} in height and recorded from \SI{60}{\centi\meter} above the water surface. Pebbles are added to the bottom of the tank to increase friction and minimize slippage caused by the smooth glass surface of the tank while crawling underwater. 

\begin{figure}[ht]
  \centering  \includegraphics[width=\linewidth]{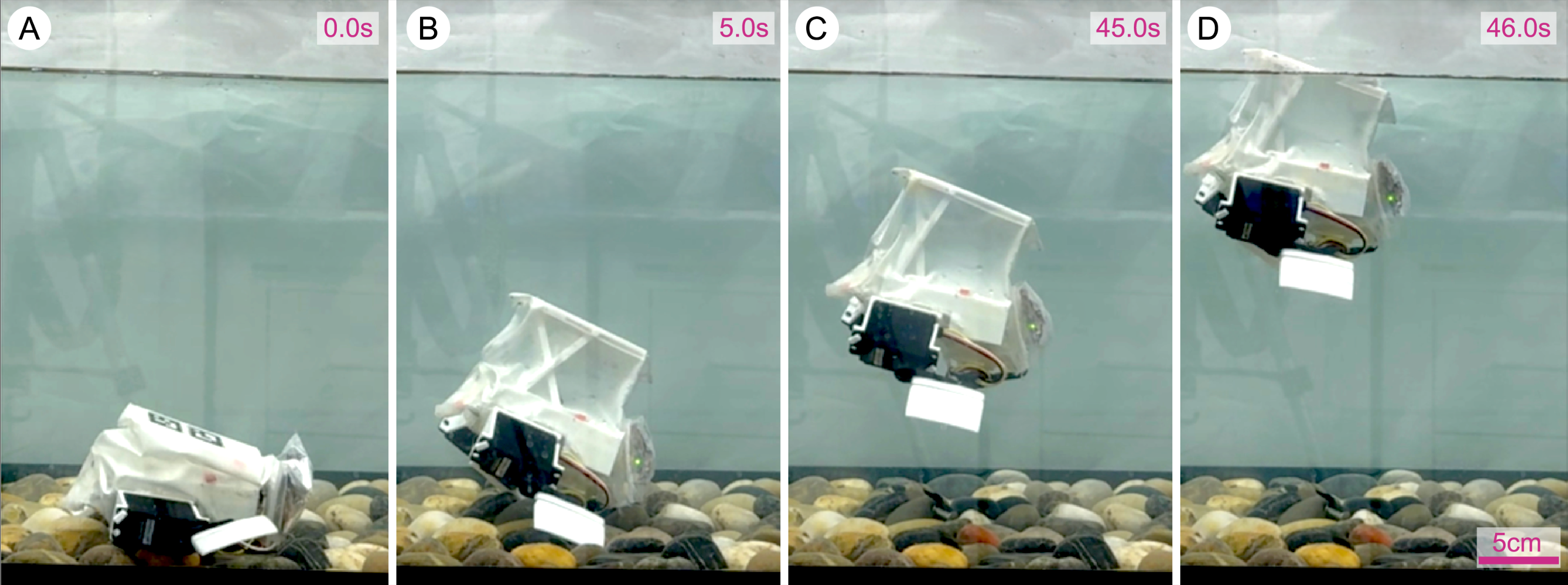}
  \caption{Sequential frames showing the robot’s transition from negative to positive buoyancy. (A) at \SI{0}{\second}, it is fully compressed and resting on the underwater floor. (B) at \SI{5}{\second}, body expansion begins and starts lifting. (C) at \SI{45}{\second}, it becomes positively buoyant and ascends. (D) by \SI{46}{\second}, it reaches the surface, completing the shape morphing driven buoyancy.}
  \label{fig:buoyant_transition}
  \vspace{-1.0em}
\end{figure}

To assess buoyancy transitions shown in Figure~\ref{fig:buoyant_transition}, ArUco markers are attached to the bottom side face of the robot. A camera is placed \SI{90}{\centi\meter} in front of the water tank to capture a side view of the robot. The robot's motion and buoyancy states are tracked using video recordings. All video data are processed using the OpenCV ArUco library to track ArUco markers. To reduce noise in the data, a low-pass filter is applied during post-processing to smooth the tracking curves. Data visualization and analysis are performed using Python’s matplotlib library.
 
To capture the robot's transition dynamics, additional experiments are conducted as shown in Figure~\ref{fig:transition}. The incline ramp at \SI{20}{\degree} is created using an acrylic sheet and secured in place from the shelf to the edge of the water tank using duct tape. The robot is placed at the top of the ramp and a forward movement command is transmitted over Wi-Fi. The robot crawls down the ramp, falls into the water and sinks to the bottom of the tank in its compressed state, which has negative buoyancy. Then it crawls along the bottom of the tank. Subsequently, a command is sent to morph the robot’s body, allowing it to transition to a positively buoyant state and float on the water surface. Once on the surface, the morphing signal is stopped and the robot is commanded to swim backward.

\subsection{Multi-environment locomotion and transition}

We illustrate the transitions of the robot between various environments in Figure~\ref{fig:transition}, beginning with land based locomotion, followed by underwater crawling, and culminating in a buoyancy-driven ascent and backward swimming on the water surface. The sequence demonstrates the robot's adaptability across diverse terrains and conditions (See supplementary video).

The robot begins its journey from the cardboard ramp that mimics land in the upper left, crawling downward along a ramp inclined at \SI{20}{\degree} for \SI{7.5}{\second}. Upon reaching the edge, the robot drops into the water. Due to its compressed body, it creates negative buoyancy and sinks to the bottom of the tank. To simulate a realistic underwater scenario, we cover the base of the tank with pebbles, and to provide additional friction for underwater locomotion.  

At the bottom, the robot crawls underwater for \SI{25}{\second} before stopping. The robot then starts to expand its body to increase the volume, thus displacing more water and achieving positive buoyancy. This buoyancy adjustment takes approximately \SI{45}{\second} plus additional \SI{10}{\second} for the user to send the command and reception by the robot. Finally, the robot reaches the water surface at \SI{90}{\second}. In the end, the robot travels backward swimming on the surface of the water for \SI{15}{\second}, completes its journey at \SI{105}{\second} and completes the operational cycle. This demonstration highlights the robot's capability for seamless transitions across land, underwater, and water surface environments, showcasing its versatility in dynamic settings. 

\begin{figure}
  \centering  \includegraphics[width=0.85\linewidth]{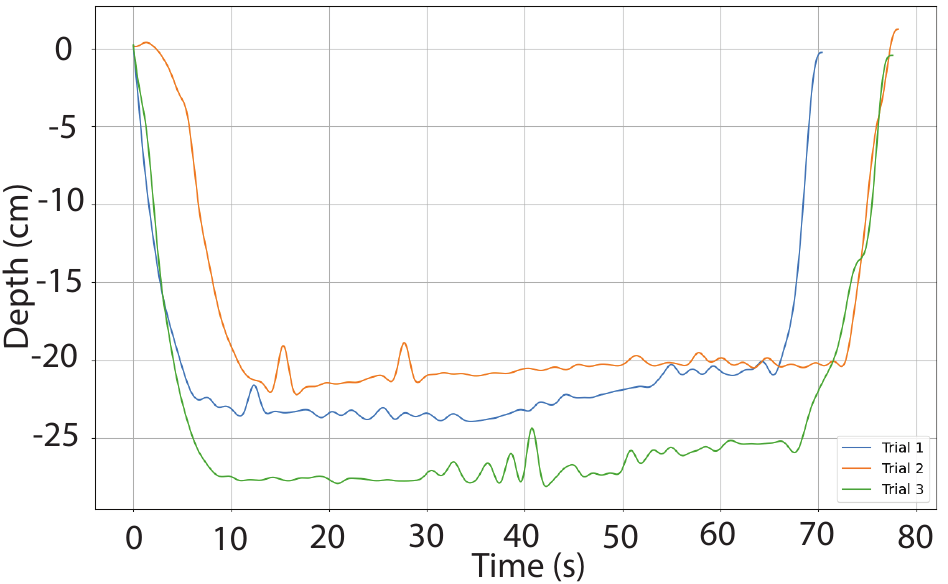}
  \caption{Depth trajectories from three trials, with the water surface defined at \SI{0}{\centi\meter} and the tank bottom at approximately \SI{-30}{\centi\meter}. Depth variation is due to pebbles at the bottom. In all trials, the robot sinks within \SI{7}{\second} and resurfaces in approximately \SI{60}{\second} via shape morphing.}
  \label{fig:depth}
  \vspace{-1.0em}
\end{figure}

\subsection{Buoyancy experiment}
To demonstrate the robot’s capability of adjusting buoyancy, we conduct experiments in a water tank, starting with the robot fully expanded and floating on the surface. The robot then gradually compresses, causing it to sink, before re-expanding to return to the surface.

Figure~\ref{fig:depth} represents the water surface at \SI{0}{\centi\meter} and the bottom of the tank at a depth of \SI{-30}{\centi\meter}. Variations in the depth of the water surface (\SI{2.5}{\centi\meter} to \SI{5}{\centi\meter}) are attributed to pebbles at the bottom. In all trials, the robot starts on the surface and is compressed to initiate sinking. On average, the robot requires approximately \SI{7}{\second} to reach the bottom. Afterward, the robot is re-expanded, taking about \SI{60}{\second} to return to the surface. 

Figure~\ref{fig:buoyant_transition} illustrates the four key stages of this process. At \SI{0}{\second} (Figure~\ref{fig:buoyant_transition}.A), the robot is fully compressed and negatively buoyant, resting at the bottom. By \SI{5}{\second} (Figure~\ref{fig:buoyant_transition}.B), the robot starts to expand, lifting its front side. At \SI{45}{\second} (Figure~\ref{fig:buoyant_transition}.C), the robot achieves significant buoyancy, rising through the water column. Finally, at \SI{46}{\second} (Figure~\ref{fig:buoyant_transition}.D), the robot reaches the surface, completing its buoyancy transition.

\subsection{Locomotion performance on land and water} 

The robot exhibits multi-environment locomotion capabilities, including crawling on land, swimming on the water surface, and crawling along the underwater floor, as demonstrated in the supplementary video. During terrestrial and underwater crawling, the robot's fins rotate sequentially in a full \(360^\circ\) cycle to generate forward motion. A proportional-integral-derivative (PID) controller is employed to monitor fin positions and maintain synchronized gait patterns. In contrast, for swimming, the fins oscillate through a \(180^\circ\) range of motion, with faster backward strokes and slower recovery strokes to achieve net forward propulsion. The robot’s locomotion was evaluated over three independent trials, with motion trajectories tracked using ArUco markers. Figure~\ref{fig:multi-environment} illustrates the multi-environment locomotion of the robot in different environments.  

Figures~\ref{fig:multi-environment}(A–C) display the trajectories of the robot in different environments, and the distance vs. time plots displayed below. In Figure~\ref{fig:multi-environment}.A, the robot is shown crawling on flat ground at \SI{0}{\second}, \SI{20}{\second}, and \SI{40}{\second}. Similarly, Figure~\ref{fig:multi-environment}.B illustrates the robot crawling underwater over pebbles at \SI{0}{\second}, \SI{60}{\second}, and \SI{120}{\second}. Lastly, Figure~\ref{fig:multi-environment}.C captures the robot swimming on the water surface at \SI{0}{\second}, \SI{20}{\second}, and \SI{36}{\second}.  

From the distance versus time plots for three trials conducted in each environment over a duration of 60 seconds, we show the consistent performance during land crawling and swimming, with an average speed of \SI{0.70}{\centi\meter\per\second} and  \SI{0.75}{\centi\meter\per\second} respectively. Although the speed of underwater crawling remains constant in different trials, underwater crawling is markedly slower, achieving an average speed of only \SI{0.24}{\centi\meter\per\second}. We attribute this to the uneven terrain caused by the pebbles.

\section{Discussion and Conclusions}

This work presented the design, modeling, and experimental validation of \robotname, an untethered, shape-morphing amphibious robot capable of multi-environment locomotion across land, underwater, and water surfaces. Unlike previous shape-morphing systems that rely on thermal actuation~\cite{baines2022multi,sun2023embedded,patel2023highly} or multiple high-power actuators~\cite{sihite2023multi}, \robotname\ employs a single linear actuator mechanically coupled to a bell-crank linkage to simultaneously change body volume and fin orientation by \SI{90}{\degree}. This coupling enables seamless transitions between locomotion modes and buoyancy states through purely geometric morphing. Experiments demonstrated crawling on land and underwater, swimming on the surface, and buoyancy-driven transitions between these modes using less than \SI{7}{\joule} of energy per morph (Table~\ref{tab:morphing_comparison}) which is orders of magnitude lower than shape memory based morphing systems.

\robotname\ demonstrates a unique combination of underwater locomotion and active buoyancy adaptation within a single untethered platform, a capability not previously reported in amphibious robots. The design leverages simple, low-power components and 3D-printed parts, providing a compact and energy-efficient system architecture suitable for long-duration operation. These results highlight the potential of mechanically coupled morphing mechanisms as a practical and scalable approach for amphibious robotics.

Despite these advantages, the current prototype has some limitations. The buoyancy control system operates in an open loop without onboard sensing, which restricts precise depth control near the neutral buoyancy point. The experiments were conducted in a controlled tank and did not account for dynamic flow effects or inclined terrain interactions. Additionally, the morphing process currently relies on ambient air inside a sealed TPU chamber, which limits rapid or precise volume regulation.

Future work will focus on improving the robot’s autonomy, control, and environmental robustness. Integrating onboard pressure sensors will enable closed-loop feedback for real time buoyancy regulation and adaptive gait transitions. Employing miniature air pumps or compressed air reservoirs could allow active pressure modulation inside the body, providing faster and finer control of buoyancy. Expanding experimental validation to larger tanks and outdoor aquatic environments will help assess performance under realistic flow and wave conditions. These improvements aim to extend \robotname’s capabilities toward autonomous, amphibious operation in natural terrains such as environmental exploration, inspection, and search-and-rescue tasks.

\bibliographystyle{ieeetr}
\bibliography{bibtex/bib}
\end{document}